\documentclass[sigconf,screen]{acmart}
\AtBeginDocument{%
  \providecommand\BibTeX{{%
    \normalfont B\kern-0.5em{\scshape i\kern-0.25em b}\kern-0.8em\TeX}}}

\copyrightyear{2021} 
\acmYear{2021} 
\setcopyright{acmlicensed}\acmConference[CHI '21]{CHI Conference on Human Factors in Computing Systems}{May 8--13, 2021}{Yokohama, Japan}
\acmBooktitle{CHI Conference on Human Factors in Computing Systems (CHI '21), May 8--13, 2021, Yokohama, Japan}
\acmPrice{15.00}
\acmDOI{10.1145/3411764.3445717}
\acmISBN{978-1-4503-8096-6/21/05}

\acmSubmissionID{9669}

\newcommand{\Comment}[1]{}

\newcommand{\etal}{\mbox{\it et al.}}

\newcommand{\eg}{\mbox{\it e.g.}}
\newcommand{\ie}{\mbox{\it i.e.}}

\newcommand{\bi}{\begin{list}{$\bullet$}{
    \setlength{\leftmargin}{1.5 em}
    \setlength{\itemsep}{0 pt}
    \setlength{\topsep}{3 pt}
    \setlength{\parsep}{3 pt}
    \setlength{\partopsep}{0 pt}
    \setlength{\labelwidth}{1 em}
    \setlength{\labelsep}{0.5 em}
    \setlength{\parskip}{0cm}  }}
\newcommand{\ei}{\end{list}}

\newcommand{\BE}{\begin{enumerate}}
\newcommand{\EE}{\end{enumerate}}

\newcommand{\initab}{                           
\begin{tabbing}
XXX \= XXXX \= \kill
}
\newcommand{\begpub}{
\begin{quotation}
\noindent
}

\newcommand{\finpub}{
\end{quotation}
}

\usepackage{paralist}
\usepackage{xspace}
\usepackage{comment}
\usepackage{subcaption}
\usepackage{enumitem}
\usepackage{multicol}
\usepackage{multirow}
\usepackage{pifont}
\newcommand{\xmark}{\ding{55}}%
\usepackage{tikz}
\usepackage[normalem]{ulem}
\definecolor{csource}{rgb}{0.01, 0.42, 0.03}
\definecolor{crep}{rgb}{0, 0, 0.7}

\definecolor{mygrey}{RGB}{190, 190, 190}
\newcommand{\greydot}[1]{{\protect\tikz\protect\draw[mygrey,fill=mygrey] (0,0) circle (.5ex); #1}}

\newcommand{\checknew}[1]{{\color{black}{#1}}}

\newcommand{\checkchi}[1]{{{#1}}}

\newcommand{\chifinal}[1]{{\color{blue}{#1}}}

\newcommand{\chifinaldel}[1]{{\color{red}{#1 }}}
\renewcommand{\chifinaldel}[1]{}

\renewcommand{\chifinal}[1]{#1}

\newcommand{\dbeer}{\emph{Beer}\xspace}

\newcommand{\damz}{\emph{Amzbook}\xspace}
\newcommand{\dlsat}{\emph{LSAT}\xspace}

\newcommand{\cHuman}{Human\xspace}

\newcommand{\esingle}{Explain-Top-1\xspace}
\newcommand{\edouble}{Explain-Top-2\xspace}
\newcommand{\eadapt}{Adaptive\xspace}

\newcommand{\cConf}{Team~(Conf)\xspace}
\newcommand{\cAISingle}{Team~(\esingle, AI)\xspace}

\newcommand{\cAIAdapt}{Team~(\eadapt, AI)\xspace}

\newcommand{\cExpertDouble}{Team~(\edouble, Expert)\xspace}
\newcommand{\cExpertAdapt}{Team~(\eadapt, Expert)\xspace}

\newcommand{\tquote}[1]{\emph{\textcolor{darkgray}{``#1''}}}

\newcommand{\squote}[1]{\emph{``#1''}}

\newcommand{\rsec}[1]{Section~}

\begin{document}

\title{Does the Whole Exceed its Parts? The Effect of AI Explanations on Complementary Team Performance}



\author{Gagan Bansal}
\authornote{Equal contribution.}
\email{bansalg@cs.washington.edu}
\author{Tongshuang Wu}
\authornotemark[1]
\email{wtshuang@cs.washington.edu}
\affiliation{%
  \institution{University of Washington}
}

\author{Joyce Zhou}
\authornote{Made especially large contributions.}
\email{jyzhou15@cs.washington.edu}
\author{Raymond Fok}
\authornotemark[2]
\email{rayfok@cs.washington.edu}
\affiliation{%
  \institution{University of Washington}
}

\author{Besmira Nushi}
\email{besmira.nushi@microsoft.com}
\affiliation{%
  \institution{Microsoft Research}
}

\author{Ece Kamar}
\email{eckamar@microsoft.com}
\affiliation{%
  \institution{Microsoft Research}
}

\author{Marco Tulio Ribeiro}
\email{marcotcr@microsoft.com}
\affiliation{%
  \institution{Microsoft Research}
}
\author{Daniel S. Weld}
\email{weld@cs.washington.edu}
\affiliation{%
  \institution{University of Washington \&}
  \institution{Allen Institute for Artificial Intelligence}
}
\renewcommand{\shortauthors}{G. Bansal et al.}


\begin{abstract}
Many researchers \checkchi{motivate} explainable AI \checkchi{with} studies showing that human-AI team performance on decision-making tasks improves when the AI {\em explains} its recommendations. 
However, prior studies observed improvements from explanations only when the AI, alone, outperformed both the human and the best team. 
Can explanations \checkchi{help lead to} {\em complementary performance}, where team  accuracy is higher than either the human or the AI working solo? 
We conduct mixed-method user studies on three datasets, where an AI with accuracy comparable to humans helps participants solve a task (explaining itself in some conditions). 
While we observed complementary improvements from AI augmentation, they were {\em not} increased by explanations. Rather, explanations increased the chance that humans will accept the AI's recommendation, regardless of its correctness. Our result poses new challenges for human-centered AI: Can we develop explanatory approaches that encourage appropriate 
trust in AI, and therefore \checkchi{help generate (or improve)} complementary performance?
\end{abstract}



\maketitle

\section{Introduction}

\definecolor{explstr}{HTML}{b51700}
\definecolor{prevstr}{HTML}{ff978d}
\definecolor{ourstr}{HTML}{56c1ff}

\begin{figure*}[ht]
 \centering
 \includegraphics[width=0.9\linewidth]{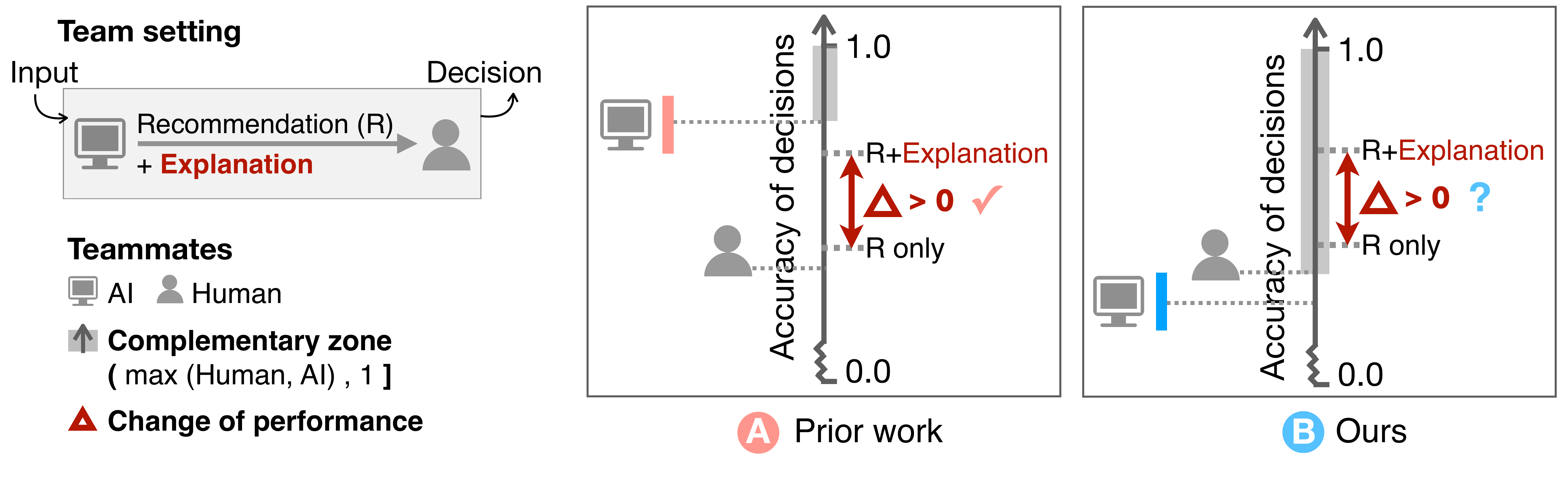}
 \vspace{-10pt}
 \caption{(Best viewed in color) Do AI explanations lead to complementary team performance? 
 In a team setting, when given an input, the human uses (usually imperfect) recommendations from an AI model to make the final decision. 
 We seek to understand if automatically generated \textcolor{explstr}{explanations} of the AI's recommendation improve team performance compared to baselines, such as simply providing the AI's recommendation, $R$, and confidence.
 (A) Most \textcolor{prevstr}{previous work} concludes that explanations improve team performance (\ie, $\Delta_A > 0$); however, it usually considers settings where AI systems are much more accurate than people and even the human-AI team.
 (B) \textcolor{ourstr}{Our study} considers settings where human and AI performance is comparable to allow room for complementary improvement. We ask, ``Do explanations help in this context, and how do they compare to simple confidence-based strategies?'' (Is $\Delta_B > 0$?). 
 }
 \Description{
 The figure compares our work with prior work.
 Most previous work concludes that explanations improve team performance; however, it usually considers settings where AI systems are much more accurate than people and even the human-AI team.
Our study considers settings where human and AI performance is comparable to allow room for complementary improvement. We ask, ``Do explanations help in this context, and how do they compare to simple confidence-based strategies?''
}

 \label{fig:landing}
\end{figure*}

Although the accuracy of Artificial Intelligence (AI) systems is rapidly improving, in many cases, it remains risky for an AI to operate autonomously, \eg, in high-stakes domains or when legal and ethical matters prohibit full autonomy.
A viable strategy for these scenarios is to form \emph{Human-AI teams}, in which the AI system augments one or more humans by recommending its predictions, but people retain agency and have accountability on the final decisions. 
Examples include AI systems that predict likely hospital readmission to assist doctors with correlated care decisions~\cite{bayati-plos14,bussone-ichi2015,caruana-kdd15,wiens-jmlr16} and AIs that estimate recidivism to help judges decide whether to grant bail to defendants~\cite{angwin-propublica16, hayashi-cscw2017}. 
\chifinaldel{In such scenarios, the human-AI team is expected to perform better than either would alone.}
\chifinal{
In such scenarios, it is important that the human-AI team achieves \emph{complementary performance} (\ie, performs better than either alone):
From a decision-theoretic perspective, a rational developer would only deploy a team if it adds utility to the decision-making process~\cite{morgenstern1953theory}. 
For example, significantly improving decision accuracy by closing deficiencies in automated reasoning with human effort, and vice versa~~\cite{horvitz2007complementary, tan2018invest}.
}

Many researchers have argued that such human-AI teams would be improved if the AI systems could {\em explain their reasoning}.
In addition to increasing trust between humans and machines or \checkchi{improving the speed of decision making}, one hopes that an explanation should help the responsible human know when to trust the AI's suggestion and when to be skeptical, \eg, when the explanation doesn't ``make sense.''
Such {\em appropriate reliance}~\cite{lee-appropriate-reliance} is crucial for users to leverage AI assistance and improve task performance~\cite{bligic-iui05}.
Indeed, at first glance, it appears that researchers have already confirmed the utility of explanations on tasks ranging from medical diagnosis~\cite{cai2019hello, lundberg-nature18}, data annotation~\cite{schmidt-aaai19} to deception detection~\cite{lai-fat19}. 
In each case, the papers show that, when the AI provides explanations, \checknew{team accuracy reaches a level higher than human-alone.}

However, a careful reading of these papers shows another commonality: 
in every situation, while explanations are shown to help raise team performance {\em closer} to that of the AI, one would achieve an even better result by stripping humans from the loop and letting the AI operate autonomously (Figure~\ref{fig:landing}A \&\ Table~\ref{table:prior_work}).
Thus, the existing work suggests several important open questions for the AI and HCI community: 
Do explanations \checkchi{help achieve} \emph{complementary performance} 
by enabling humans to anticipate when the AI is potentially incorrect?
Furthermore, do explanations provide significant value over simpler strategies such as displaying the AI's uncertainty? In the quest to build the best human-machine \checkchi{teams}, such questions deserve critical attention.

\chifinaldel{
To explore these questions, we conduct new studies where we control the study design such that the AI's accuracy is \emph{comparable} to the human's (Figure~\ref{fig:landing}B). 
This decision simulates a setting where there is a stronger incentive to deploy human-AI teams, \eg, because \checkchi{there exists more potential for complementary performance (by correcting each other's mistakes), and where simple heuristics such as directly following the AI are unlikely to achieve the highest performance.}

We are especially interested in high-stakes tasks, such as medical diagnosis and recidivism prediction.
However, since conducting controlled studies on these domains is challenging and ethically fraught, we started by selecting three common-sense tasks which can be tackled by crowd workers with little training: 
a set of \dlsat questions that require logical reasoning, and sentiment analysis of book and beer reviews. 
We measure human skill on these problems and then control AI accuracy by purposely selecting study samples where AI has comparable accuracy.
Among AI-assisted conditions that displayed explanations, we varied the strategies of explanations by explaining just the predicted class versus explaining the top-two classes. 
We also introduced a novel {\em Adaptive Explanation} mechanism, which switches explanation strategies based on the AI's confidence to discourage blind trust and achieve appropriate reliance. 
Since explanation {\em quality} is also a likely factor in its effectiveness and AI technology keeps improving, we also varied the source of explanations and considered those generated by state-of-the-art AI methods as well as by skilled humans.

In summary, through both qualitative and quantitative analyses, our paper makes the following contributions:

\begin{compactitem}
 \item 
 We highlight an important limitation of previous work on explainable AI. 
 While several studies have shown that explanations of AI predictions can increase team performance (Table~\ref{table:prior_work}), these results were only shown for settings where the AI system was significantly more accurate than the human partner and the human-AI team. To our knowledge, no previous work has evaluated the effectiveness of AI explanations at helping produce (or improve) {\em complementary performance}, where the human-AI team outperforms both solo human and AI. 
 
 \item
 We develop an experimental setting to study human-AI complementary performance 
 and conducted studies with 1626 users on three tasks, using a variety of explanation sources and strategies. 
 We observed complementary performance on every task, but surprisingly, explanations did {\em not} improve team performance compared to simply showing the AI's confidence.
 Explanations increased team performance when the AI system was correct, but {\em decreased} it when the AI erred, \checkchi{indicating an increased likelihood of human agreement with the AI even for incorrect predictions} --- so, the net change in performance was minimal
 
 \item Our novel {\em \eadapt Explanations} method, which attempts to solve the problem of blind agreement by reducing human trust when the AI has low confidence, failed to produce significant improvement in final team performance over other explanation types. However, there is suggestive evidence that the adaptive approach is promising for high-quality explanations. 
 Through extensive qualitative analysis, we extract potential causes for the behavioral differences among datasets and summarize them as design implications.
\end{compactitem}
}

\chifinal{
To explore these questions, we conduct new experiments where we control the study design, ensuring that the AI's accuracy is \emph{comparable} to the human's (Figure~\ref{fig:landing}B). 
Specifically, we measure the human skill on our experiment tasks and then control AI accuracy by purposely selecting study samples where AI has comparable accuracy.
This setting simulates situations where there is a strong incentive to deploy human-AI teams, \eg, because there exists more potential for complementary performance (by correcting each other's mistakes), and where simple heuristics such as blindly following the AI are unlikely to achieve the highest performance.

We selected three common-sense tasks that can be tackled by crowd workers with little training: sentiment analysis of book and beer reviews and a set of \dlsat questions that require logical reasoning.
We conducted large-scale studies using a variety of explanation sources (AI versus expert-generated) and strategies (explaining just the predicted class, or explaining other classes as well). 
We observed complementary performance on every task, but --- surprisingly --- explanations did not appear to offer benefit compared to simply displaying the AI's confidence. Notably, explanations increased reliance on recommendations even when the AI was incorrect.
Our result echoes prior work on inappropriate trust on systems~\cite{kaur-chi2020, mitchell2019model}, \ie, explanations can lead humans to either follow incorrect AI suggestions or ignore the correct ones~\cite{bussone-ichi2015, stumpf2009interacting}.
However, using end-to-end studies, we go one step further to quantify the impact of such over-reliance on objective metrics of team performance.

As a first attempt to tackle the problem of blind reliance on AI, we introduce {\em Adaptive Explanation}.
Our mechanism tries to reduce human trust when the AI has low confidence: it only explains the predicted class when the AI is confident, but also explains the alternative otherwise.
While it failed to produce significant improvement in final team performance over other explanation types, there is suggestive evidence that the adaptive approach can push the agreement between AI predictions and human decisions towards the desired direction.

Through extensive qualitative analysis, we also summarize potential factors that should be considered in experimental settings for studying human-AI complementary performance.
For example, the difference in expertise between human and AI affects whether (or how much) AI assistance will help achieve complementary performance, and the display of the explanation may affect the human's collaboration strategy. In summary:

}

\chifinal{
\begin{compactitem}
 \item 
 We highlight an important limitation of previous work on explainable AI: While many studies show that explaining predictions of AI increases team performance (Table~\ref{table:prior_work}), they all consider cases where the AI system is significantly more accurate than both the human partner and the human-AI team. In response, we argue that AI explanations for decision-making should aim for complementary performance, where the human-AI team outperforms both solo human and AI.

 \item
 To study complementary performance, we develop a new experimental setup and use it in studies with 1626 users on three tasks\footnote{\chifinal{All the task examples and the collected experiment data are available at \url{https://github.com/uw-hai/Complementary-Performance}.}} to evaluate a variety of explanation sources and strategies. We observe complementary performance in every human-AI teaming condition.
 
 \item However, surprisingly, we do not observe any significant increase in team performance by communicating explanations, compared to simply showing the AI's confidence. 
 Explanations often increased accuracy when the AI system was correct but, worryingly, {\em decreased} it when the AI erred, resulting in a minimal net change --- even for our adaptive explanations.
 Through qualitative analysis, we discuss potential causes for failure of explanations, behavioral differences among tasks, and suggest directions for developing more effective AI explanations.
\end{compactitem}
}

 

\begin{table*}
 \centering
 \setlength{\tabcolsep}{5pt}
 \begin{tabular}{ c c c c c c c}
 \toprule
 
 \multirow{2}{*}{\textbf{Domain}} & \multirow{2}{*}{\textbf{Task}} &\multicolumn{5}{c}{\textbf{Performance}} \\
 \cmidrule(lr){3-7}
 & & {Metric} & {$\!\!\!\!\!$Human alone} & {AI alone} & {Team} & {Complementary?} \\
 \midrule
 \multirow{5}{*}{Classification$\!$}
 & Deceptive review~\cite{lai-fat19} & Accuracy $\uparrow$ & 51.1\% & \textbf{87.0\%} & 74.6\% & \xmark \\
 & Deceptive review~\cite{lai-chi2020} & Accuracy $\uparrow$ & \checkchi{54.6}\% & \textbf{86.3\%} & 74.0\% & \xmark \\
 & Income category~\cite{zhang-arxiv2020} & Accuracy $\uparrow$ & 65\% & \textbf{75\%} & 73\% & \xmark \\
 & Loan defaults~\cite{green-cscw2019} & Norm. Brier $\uparrow\!\!\!$ & 0 & \textbf{1} & 0.682 & \xmark \\
 & Hypoxemia risk~\cite{lundberg-nature18} & AUC $\uparrow$ & 0.66 & \textbf{0.81} & 0.78 & \xmark \\
 & \checkchi{Nutrition prediction~\cite{buccinca-iui20}} & \checkchi{Accuracy $\uparrow$} & \checkchi{0.46} & \checkchi{\textbf{0.75}} & \checkchi{0.74} & \xmark \\
 \midrule
 QA & Quiz bowl~\cite{feng-iui19} & \multicolumn{4}{c}{\tquote{AI outperforms top trivia players.}} & \xmark\\
 \midrule
 Regression & House price~\cite{sangdeh-arxiv18} & Avg. Absolute Error $\downarrow$ & \$331k & \textbf{\$200k} & \$232k & \xmark \\
 \bottomrule
 \end{tabular}
 \caption{
 Recent studies that evaluate the effect of automatically generated explanations on human-AI team performance. While explanations did improve team accuracy, the performance was not complementary --- acting autonomously, the AI would have performed even better.
 For papers with multiple domains or experiments, we took one sample with the most comparable human and AI performance.
 \checkchi{$\uparrow$ (or $\downarrow$) indicates whether the metric should be maximized (or minimized).}
 }
  \Description{
 The table summarizes recent studies that evaluate the effect of automatically generated explanations on human-AI team performance. 
 While explanations did improve team accuracy, the performance was not complementary --- acting autonomously, the AI would have performed even better.
}
 \label{table:prior_work}
 \vspace{-10pt}
\end{table*}

\section{Background and Related Work}
\label{sec:background}

Explanations can be useful in many scenarios where a human and AI interact: \checkchi{transparently} communicating model predictions \cite{ribeiro-kdd16, feng-iui19, koh-icml17,kaur-chi2020,bligic-iui05}, teaching humans tasks like translation~\cite{glassman2015overcode, mac2018teaching} or content moderation~\cite{jhaver2019does}, augmenting human analysis procedure~\cite{jhaver2019does} or creativity~\cite{clark2018creative}, legal imperatives~\cite{miller-explanation17,weld-cacm19}, etc.
Various studies have evaluated the effect of explanations from different dimensions, including whether the explanation improves users' trust in the AI~\cite{yu2019trust, kunkel2019let} or enables users to simulate the model predictions~\cite{sangdeh-arxiv18,chandrasekaran-emnlp18}, or assists developers to debug models~\cite{bhatt-fat2020,kaur-chi2020}. 

In this paper, we focus explicitly on {\em AI-assisted decision making} scenarios~\cite{bansal-aaai19, wang2019designing}, where an AI assistant (\eg, a classification model) makes recommendations to a human (\eg, a judge), who is responsible for making final decisions (\eg, whether or not to grant bail).
In particular, we assess performance in terms of the \emph{accuracy} of the human-AI team. 
\checkchi{While other metrics can be used for evaluation (more discussed in Section~\ref{subsec:limitations}), we directly evaluate end-to-end team accuracy for three reasons.}
First, deploying such a human-AI team is ideal if it achieves {\em complementary performance}, \ie, if it outperforms both the AI and the human acting alone.
\checkchi{Second, evaluating explanations using proxy tasks (such as whether humans can use it to guess the model's prediction) can lead to different, misleading conclusions for achieving best team performance than an end-to-end evaluation~\cite{buccinca-iui20}.} 
\checkchi{Third, } AI-assisted decision making is often listed as a major motivation for AI explanations. In recent years numerous papers have employed user studies to show that human accuracy increases if the AI system explains its reasoning for tasks as diverse as medical diagnosis, predicting loan defaults, and answering trivia questions.
However, as summarized in Table \ref{table:prior_work}, complementary performance was not observed in any of these studies -- in each case, adding the human to the loop {\em decreased} performance compared to if AI had acted alone.

For example, in Lai~\etal~\cite{lai-chi2020, lai-fat19}, MTurk workers classified deceptive hotel reviews with predictions from SVM and BERT-based models, as well as explanations in the form of inline-highlights. 
However, models outperformed every team (see Table~1 and Figure~6 in \cite{lai-chi2020}). 
Zhang~\etal~\cite{zhang-arxiv2020} noticed the superior behavior of the models in Lai~\etal's work, and evaluated the accuracy and trust calibration where the gap between human and the AI performances was less severe.
Still, on their task of income category prediction, their Gradient Boosted Trees model had 10\% higher accuracy compared to their MTurk workers, which seemed borderline ``comparable'' at best. 
Furthermore, when run autonomously, their AI model performed just slightly better than the best team (see Section~4.2.2 and Figure~10 in \cite{zhang-arxiv2020}). 
A similar performance trend is observed on tasks other than classification.
In Sangdeh~\etal~\cite{sangdeh-arxiv18}, MTurk workers predicted house price using various regression models that generated explanations in terms of most salient features. Their models' predictions resulted in lowest error (See Figure~6 in \cite{sangdeh-arxiv18}). 
In Feng~\etal~\cite{feng-iui19}, experts and novices played Quiz Bowl with recommendation from Elastic Search system. 
The system explained its predictions by presenting training examples that were influential, and using inline-highlights to explain the connection between question and evidence. 
However, Feng~\etal~do not report the exact performance of the AI on their study sample, but mention its superiority in Section 3.1 in \cite{feng-iui19} pointing out that it outperforms top trivia players.
One possible exception is Bligic \& Mooney (2005)~\cite{bligic-iui05}, who probably achieved complementary performance on their task of recommending books to users. However, they did not compare explanations against simple baselines, such as showing the book title or the system confidence (rating). 

At least two potential causes account for the absence of complementary performance in these cases. 
First, task design may have hindered collaboration: previous researchers considered AI systems whose accuracy was substantially higher than the human's, leading to a small zone with potential for complementary performance (see Figure~\ref{fig:landing}).
For example, this may have made it more likely that human errors were a superset of the AI's, reducing the possibility of a human overseer spotting a machine mistake.
Second, even when the task has the potential for complementary performance, it is unclear if the collaboration mechanisms under study supported it. 
Collaboration factors like incentives, the format of explanations, and whether AI's uncertainty was displayed may drive the human towards simple, less collaborative heuristics, such as ``always trust the AI'' or ``never trust the AI.''

\begin{table}[tb]
\setlength{\tabcolsep}{10pt}
\small
 \centering
 \begin{tabular}{l l l}
 \toprule
 Explain. Strategies & Explain. Sources & Tasks \\
 \cmidrule(lr){1-1}\cmidrule(lr){2-2}\cmidrule(lr){3-3}
 \esingle\ \greydot & AI \greydot & \dbeer\ \greydot{} \\
 \edouble\ \greydot & Expert & \damz \\
 \eadapt & & \dlsat \\
 \bottomrule
 \end{tabular}
 \caption{\chifinal{An overview of our tasks, explanation strategies and sources.
 We ran our pilot studies  (Section~\ref{subsec:pilot}) with conditions marked with \greydot. Based on the pilot results, we added adaptive explanations and expert explanations (Section~\ref{sec:conditions_sentiment}). Along with two additional domains, these form  the conditions for our final study conditions (Section~\ref{subsec:hypothesis_condition}).}}
 
\Description{
This figure provides an overview of the tasks, explanation strategies and sources used.
}
 \label{table:exp_meta}
 \vspace{-15pt}
\end{table}

\section{Setup and Pilot Studies}

\chifinal{
To better understand the role of explanations in producing complementary performance, we enlarge the zone of potential complementarity by matching AI accuracy to that of an average human,\footnote{Of course, complementary performance may be possible even in situations when one of the team partners is significantly more accurate than the other. 
For example, a low-accuracy team member may be valuable if their errors are independent, because they may be able to spot mistakes made by the team majority. 
However, it is more difficult to observe complementary performance in such settings, so we first consider the case where humans and AI have similar accuracy.
If explanations cannot provide value in such settings, it will be even more difficult to show complementary performance when teammates have disparate skills.}
and investigate multiple explanation styles on several domains 
(Section~\ref{sec:domains}).
As Table~\ref{table:exp_meta} summarizes, we first designed and conducted pilots studies~(Sections~\ref{subsec:pilot}) and used them to inform our final study and hypotheses~(Section~\ref{sec:formal_setup}).
}

\subsection{Choice of Tasks and Explanations}
\label{sec:domains}
Since our interest is in \emph{AI-assisted decision making}, we studied the effect of {\em local} explanations on team performance -- that is, explaining each individual recommendation made by a model~\cite{ribeiro-kdd16}.
This contrasts with providing a global understanding of the full model all at once (\eg, ~\cite{lakkaraju2017interpretable}).

\chifinal{We conducted experiments on two types of tasks: text classification (sentiment analysis) and question answering.}
\chifinaldel{We are motivated by high-stakes decisions, such as medical diagnosis, but were daunted by the difficulty of running a large-scale randomized control trial in such a domain. 
Instead, we conducted studies on text classification,} 
\chifinal{Text classification} 
because it is a popular task in natural language processing (NLP) that has been used in several previous studies on human-AI teaming~\cite{feng-iui19, lai-chi2020, lipton-icmlwhi16, nguyen2018comparing, zhang-arxiv2020, Hase2020EvaluatingEA} and because it
requires little domain expertise, and is thus amenable to crowdsourcing. 
Specifically, we selected two sentiment analysis datasets \chifinal{to improve the generalization of our results}: beer reviews~\cite{beer-aij95} and book reviews~\cite{he-www16}.
More details about these datasets are in Section~\ref{subsec:dataset_sentiment}.
\checkchi{While there exist various local explanation approaches for text classification, we rely on {\em local saliency explanations}, which explain a single prediction in terms of the importance of input features (\eg, each word) towards the model's prediction (\eg, positive or negative sentiment).
} 

As commonly practiced in previous work~\cite{lai-fat19,lai-chi2020,feng-iui19},
we display explanations with {\em inline-highlights}, \ie, directly highlighting the explanation in the input text, so the user need not go back and forth between input and the explanation. 
While there exist other explanatory approaches, such as feature-importance visualization~\cite{lundberg-nature18, green-cscw2019, weerts-arxiv2019, narayanan-arxiv18, binns-arxiv2018} (more suitable for tabular data) or communicating influential training examples~\cite{yang2020visual,mac2018teaching,koh-icml17} (more suitable for images), these techniques are not ideal for text because they add an additional cognitive cost to mapping the explanation to the respective text. 
Figure~\ref{fig:ui_text} shows one example beer review.

\chifinal{
We also experimented with Law School Admission Test (LSAT) questions\footnote{\url{https://en.wikipedia.org/wiki/Law School Admission Test}} because it is more challenging. In this task,
}
every question contains four options with a unique correct answer (Figure~\ref{fig:ui_lsat}). Again, answering \dlsat questions requires no specialized knowledge except common-sense reasoning skills, such as recognizing logical connections and conflicts between arguments~\cite{yu2020reclor}. 
Because in this case it is unclear how inline-highlights could be used to communicate logical constructs (\eg, contradiction may not be visible by highlighting the input alone), we turned to narrative explanations which justify a candidate answer in natural language. \checkchi{We explain these in more detail in Section~\ref{subsec:dataset_sentiment}}.

\subsection{Pilot Study on Sentiment Classification}
\label{subsec:pilot}

\checkchi{To iterate on the hypotheses and the associated explanation conditions for our main study (detailed later in Section~\ref{sec:formal_setup}), we conducted a pilot study on one of our datasets (\dbeer).}
\chifinaldel{The pilot study was between-subjects and involved 150 crowd workers (50 per condition) each of whom judged sentiment of 50 beer reviews using assistance from a logistic regression classifier.
The model presented its prediction and confidence in all conditions.}
\chifinal{The between-subject pilot study asked crowdworkers to judge the sentiment of 50 beer reviews with assistance from a logistic regression classifier in three conditions, each condition with 50 workers.
One condition \emph{only} showed the model prediction and confidence; the other two also included the following common {\bf explanation} strategies\footnote{the saliency scores were based on feature weights learned by the linear model~\cite{green-cscw2019, lai-chi2020}}:}

\chifinaldel{
\noindent{{\bf Explanations}}
Some conditions also included common explanation strategies based on saliency scores. 
The scores were based on feature weights learned by the linear model~\cite{green-cscw2019, lai-chi2020}:}
\begin{compactenum}
\item {\em \esingle} explains just the predicted class by highlighting the most influential words for that class.
\item {\em \edouble} explains the top two predicted classes, and unlike \esingle, it also color codes and highlights words for the other sentiment class. 
\end{compactenum}

\chifinal{
The two strategies closely align with the design in prior work~\cite{wang-etal-2016-attention,Lin2017ASS,Ghaeini-abs-1808-03894,lai-fat19}, and have been shown to be beneficial (Table~\ref{table:prior_work}).
\edouble also corresponds to Wang \etal's suggestion to mitigate heuristic biases by explaining ``multiple outcomes''~\cite{wang2019designing}.}

\noindent{{\bf Observations}}
\checkchi{
We summarize our findings from the pilot study:
\chifinaldel{
Because we made similar observations in our main study, we defer detailed discussions and implications of these observations to Section~\ref{subsec:result_quan} and Figure~\ref{fig:result}.}
\begin{compactenum}
\item Contrary to many prior works, we observed \emph{no significant changes or improvements in aggregated team accuracy by displaying either type of explanations. }
\item That said, {\em explaining just the predicted class (\esingle) performed better than explaining both (\edouble)}.
\item We also observed that \emph{explanations increased reliance on recommendations even when they were incorrect}:
explaining the predicted class slightly improved performance (compared to confidence only) when the recommendation was correct but decreased performance when it was incorrect.
\item This effect was less pronounced in \edouble, presumably because it \emph{encouraged users to consider alternatives and hence deterred over-reliance. }
In Figure~\ref{fig:ui_text}, for example, if counter-argument (d) was not highlighted, participants could easily stop reading at the highlighted first sentence and overlook the negative ending. 
\item Finally, \emph{participants indicated that they wanted higher quality explanations.} Crowd-workers were confused when explanations did not seem to correlate with model behavior.
\end{compactenum}}
\chifinal{
Because we made similar observations in our main study, we defer detailed discussions and implications of these observations to Section~\ref{subsec:result_quan} and Figure~\ref{fig:result}.
}

\subsection{Additional Explanation Strategies/Sources}
\label{sec:conditions_sentiment}

{\bf \checkchi{Added Strategy:} \eadapt Explanations.} 
The pilot study indicated that {\edouble} was more beneficial than {\esingle} when the classifier made mistakes, but not otherwise. Relying on the commonly seen correlations between mistakes and low-confidence \cite{hendrycks2016baseline}, we developed a new dynamic strategy, {\em adaptive explanation}, that switches between {\esingle} and {\edouble} depending on the AI's confidence.
This method explains the top-two classes only when the classifier confidence is below a \checkchi{task- and model-specific} threshold (described later in Section~\ref{subsec:dataset_sentiment}), explaining only the top prediction otherwise. 
Intuitively, it was inspired by an efficient assistant that divulges more information (confessing doubts and arguing for alternatives) only when it is unsure about its recommendation.
\chifinal{
Adaptive explanations can also be viewed as changing explanation according to {\em context}~\cite{abowd-springer99}.
While we limit the our context to the AI's confidence, in general, one could rely on more features of the human-AI team, such as the user, location, or time~\cite{Horvitz1999PrinciplesOM,lim-ch09}.
}
\chifinaldel{We test all three strategies in our final study\checkchi{ (detailed conditions in Section~\ref{subsec:hypothesis_condition})}. }


{\bf \checkchi{Added Source:} Expert-Generated Explanations.} 
Users in our pilot study were confused when the explanations did not make intuitive sense, 
\chifinal{perhaps due to either the quality of the underlying linear model-based AI.
While we test state-of-the-art models in the final study, 
}
\chifinaldel{
We compared a wide variety of NLP models and (saliency-based) explanation algorithms to find the most suitable AI-generated explanations (details in Appendix~\ref{subsec:explainer_select}).
}
\chifinal{
we also added {\em expert-generated} explanations to serve as an upper bound on explanation quality. We describe their annotation process in Section~\ref{subsec:dataset_sentiment}.
}

\chifinaldel{
For sentiment classification, one author created expert explanations by selecting one short, convincing text phrase span for each class (positive or negative). 
For \dlsat, we found no automated method that could generate reasonable explanations (unsurprising, given that explanations rely on prior knowledge and complex reasoning); 
instead, we used expert explanations exclusively. 
The prep book from which we obtained our study sample contained expert-written explanations for the correct answer, which one author condensed to a maximum of two sentences.
Since the book did not provide explanations for alternative choices, we created these by manually crafting a logical supporting argument for each choice that adhered to the tone and level of conciseness of the other explanations.}

\section{Final Study}
\label{sec:formal_setup}
\chifinal{
Based on our pilot studies, we formulated our final hypotheses and used them to inform our final conditions and their interface (Section~\ref{subsec:hypothesis_condition}). 
We then tested these hypotheses for several tasks and AI systems (Section~\ref{subsec:dataset_sentiment}) through crowdsourcing studies~(Section~\ref{subsec:study_procedure}).
}

\subsection{Hypotheses, Conditions, and Interface}
\label{subsec:hypothesis_condition}

\chifinaldel{Informed by our pilot study (Section~\ref{subsec:pilot}),}
We formulated the following hypotheses for sentiment analysis:

\begin{enumerate}[label=\textbf{H\arabic*}]
 {\em
 \item 
 \label{h:single_double}
 Among current explanation strategies, explaining the predicted class will perform better than explaining both classes.
 \item 
 \label{h:conf}
 The better strategies, \esingle, will still perform similarly to simply showing confidence.
 \item 
 \label{h:present}
 Our proposed \eadapt explanations, which combines benefits of existing strategies, will improve performance.
 \item
 \label{h:quality}
 \eadapt explanations would perform even better if AI could generate higher quality explanations.}
 \end{enumerate}

Since generating AI explanations for \dlsat was not feasible (Section~\ref{sec:conditions_sentiment}), we slightly modified the hypothesis for \dlsat: we omitted the hypothesis on explanation quality (\ref{h:quality}) and tested the first three hypotheses using expert- rather than AI-generated explanations.

\paragraph{\noindent {\bf Conditions}}
For both domains, we ran two baseline conditions: unassisted users (\textbf{\cHuman}), as well as a simple AI assistance that shows the AI's recommendation and confidence but \emph{no explanation} (\textbf{\cConf}).
\checkchi{We use this simple assistance because it serves as a stronger and broadly acknowledged baseline than the alternative, \ie, displaying AI's recommendation \emph{without} confidence. 
First, most ML models can generate confidence scores that, in practice, correlate with the model's true likelihood to err~\cite{hendrycks2016baseline}.
Second, displaying uncertainty in predictions can help users can make more optimal decisions~\cite{joslyn-psych13,dong-psych12,fernandes-chi18,nadav-cog09,gkatzia-arxiv16}. \chifinaldel{As a result, in high-stakes decision making it seems unethical to purposely hide this known information about imperfect AI from its users.}
Hence, we focus on evaluating whether the explanations provide {\em additional} value when shown alongside confidence scores.}
In the rest of the paper, we indicate the explanation conditions using the following template: {\bf Team (Strategy, Source)}. For example, \cAISingle indicates the condition that shows the AI's explanations for the top prediction.

\begin{figure*}[t]
 \centering
 \includegraphics[width=0.95\linewidth]{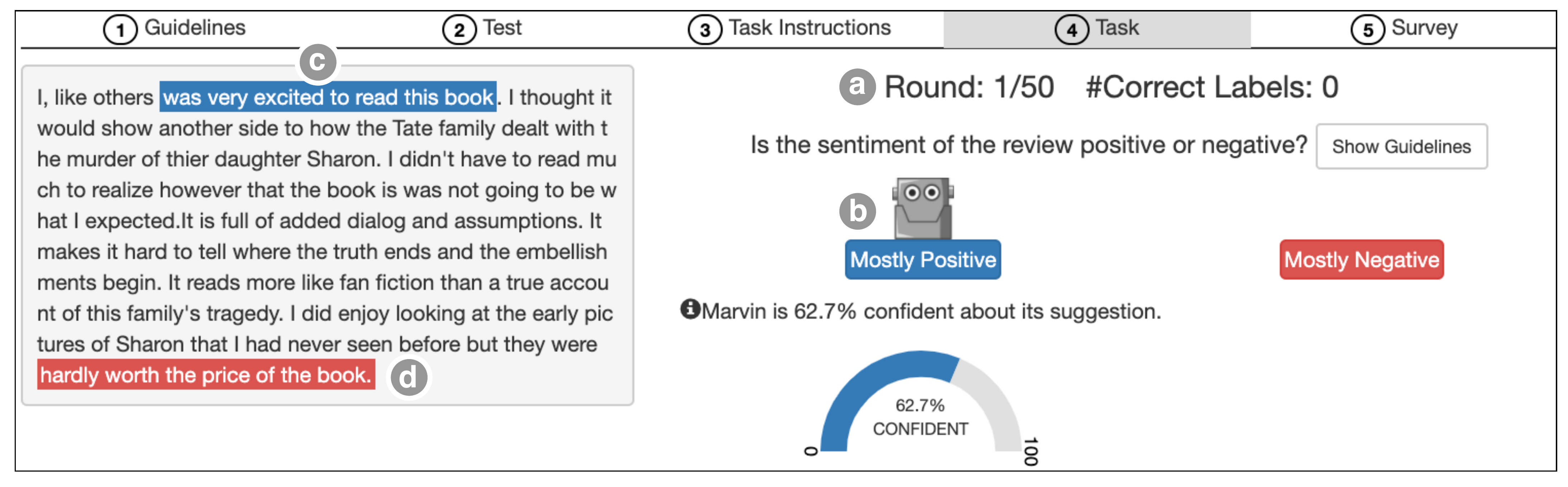}
 \vspace{-5px}
 \caption{
 A screenshot of the {\bf \cExpertAdapt} condition for the \damz reviews dataset.
 Participants read the review (left pane) and used the buttons (right pane) to decide if the review was mostly \emph{positive} or \emph{negative}. 
 The right pane also shows progress and accuracy (a). 
 To make a recommendation, the AI (called ``Marvin'') hovers above a button (b) and displays the confidence score under the button. 
 In this case, the AI incorrectly recommended that this review was positive, with confidence 62.7\%. As part of the explanation, the AI highlighted the most positive sentence (c) in the same color as the \emph{positive} button.
 Because confidence was low, the AI also highlights the most negative sentence (d) to provide a counter-argument.}
 \Description{
This figure shows a screenshot of the sentiment analysis study.
}
 \label{fig:ui_text}
 
\bigskip
 \includegraphics[width=0.95\textwidth]{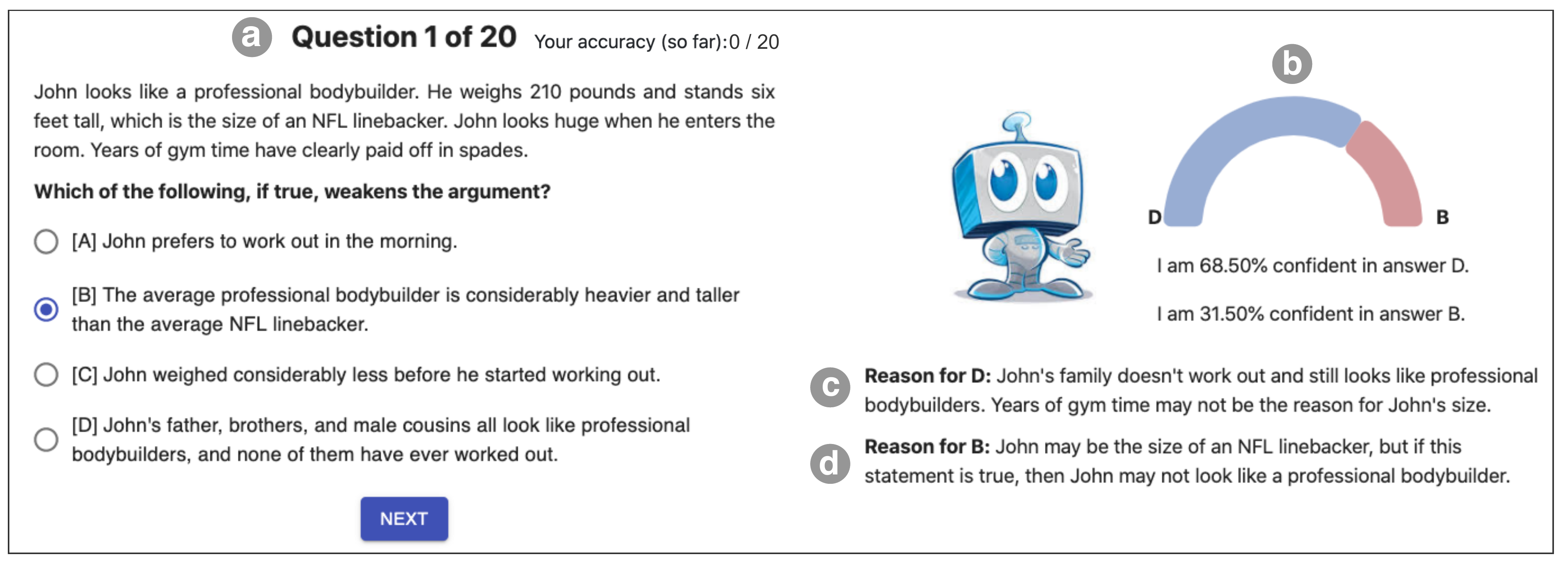}
 \vspace{-10px}
 \caption{
 A screenshot of {\bf \cExpertAdapt} for \dlsat. 
 Similar to Figure~\ref{fig:ui_text}, the interface contained a progress indicator (a), AI recommendation (b), and explanations for the top-2 predictions (c and d).
 To discourage participants from blindly following the AI, all AI information is displayed on the right.
 In (b), the confidence score is scaled so those for top-2 classes sum to 100\%.}
\Description{
This figure shows a screenshot of the LSAT study.
}
 \vspace{-5px}
 \label{fig:ui_lsat}
\end{figure*}

\paragraph{\noindent {\bf Interface}}
Figure~\ref{fig:ui_text} shows an example UI for sentiment classification for \cAIAdapt.
In all explanation conditions, explanations are displayed as inline highlights, with the background color aligned with the positive/negative label buttons.
The highlight varies by condition, \eg,
\cAIAdapt has a similar display to Figure~\ref{fig:ui_text}, except that the AI picks multiple short phrases, instead of a full sentence.
In \cAISingle the counter-argument (d) is always missing, and in \cConf no explanations are highlighted.
Figure~\ref{fig:ui_lsat} shows a screenshot of the user interface for \dlsat in the \cExpertAdapt condition.

\subsection{AI Model, Study Samples and Explanations}
\label{subsec:dataset_sentiment}

\subsubsection{Sentiment Classification}
\mbox{}\\
\textbf{Training data.}
To prepare each dataset (\dbeer and \damz) for training classification models, we binarized the target labels, split the dataset into training and test sets (80/20 split), removed class imbalance from the train split by oversampling the minority class, and further split the training set to create a validation set.

\textbf{AI Model.}
For each dataset, we fine-tuned a RoBERTa-based~\cite{Liu2019RoBERTaAR} text classifier from AllenNLP\footnote{\url{https://demo.allennlp.org/sentiment-analysis}} on the training dataset and performed hyper-parameter selection on the validation set.

\textbf{Task examples.}
For each domain, we selected 50 examples from the test set to create our study sample.
We first conducted additional pilot studies to establish the accuracy of unassisted users, which we observed were 87\% for \dbeer and 85\% for \damz\footnote{Again, each condition containing 50 crowd-workers. We estimated the human accuracy on all the three datasets with another 150 crowd-workers.}
We then selected 50 unambiguous examples so that the AI's accuracy was 84\% (\ie, comparable to human accuracy), with equal false positive and false negative rates.
\chifinal{The filtering was for keeping the task objective: If the ground-truth answer was unclear, one cannot compute or compare the accuracy of decisions.}

\textbf{Explanations.}
\checkchi{To generate saliency explanations, we used LIME, which is a popular post hoc method~\cite{ribeiro-kdd16}.
We chose this setup because the combination of RoBERTa and LIME was consistently ranked the highest among the various systems we tried in an explainer comparison study \checkchi{with human judges} (details in Appendix).} 
Despite offering accurate predictions, RoBERTa generated poorly calibrated confidence scores, a common issue with neural networks \cite{guo2017calibration}, which we mitigated with 
{\em post hoc calibration} (isotonic regression \cite{barlow1972isotonic}) on the validation set.

In particular, for \eadapt explanation, we used the classifier's median confidence as the threshold to have an equal number of 25 examples displayed as \esingle and \edouble, respectively.
The thresholds were 89.2\% for \dbeer and 88.9\% for \damz. 
We happened to explain 18 correctly predicted and 7 incorrectly predicted examples with \edouble for both datasets (leaving 1 incorrect and 24 correct cases with \esingle).
While one might learn a better threshold from the data, we leave that to future work.
\chifinal{As for expert-generated explanations, one author created expert explanations by selecting one short, convincing text phrase span for each class (positive or negative).}

\subsubsection{LSAT}
\mbox{}\\
\noindent\textbf{AI Model.}
We finetuned a RoBERTa model\footnote{\chifinal{Based on the opensource implementation: \url{https://github.com/yuweihao/reclor}.}} on ReClor~\cite{yu2020reclor}, a logic-reasoning dataset that contains questions from standardized exams like the \dlsat and GMAT.\footnote{\url{https://en.wikipedia.org/wiki/Graduate Management Admission Test}} 

\textbf{Task examples.}
We selected 20 examples from an \dlsat prep book~\cite{lsatprebook2016}.
We verified that our questions were not easily searchable online and were not included in the training dataset. 
We selected fewer \dlsat questions than for sentiment analysis, because they are more time consuming to answer and could fatigue participants:
\dlsat questions took around a minute to answer, compared to around 17 seconds for \dbeer and \damz.
The RoBERTa model achieved 65\% accuracy on these examples, comparable to the 67\% human accuracy that we observed in our pilot study.

\textbf{Explanations.}
\chifinal{We found no automated method that could generate reasonable explanations (unsurprising, given that explanations rely on prior knowledge and complex reasoning); 
instead, we used expert explanations exclusively, which is again based on the prep book.
The book contains explanations for the correct answer, which one author condensed to a maximum of two sentences.
Since the book did not provide explanations for alternative choices, we created these by manually crafting a logical supporting argument for each choice that adhered to the tone and level of conciseness of the other explanations.
\chifinal{Experts only generated explanations and did not determine the model predictions or its uncertainties.}}

\subsection{Study Procedure}
\label{subsec:study_procedure}
\paragraph{\noindent {\bf Sentiment Classification}}
For the final study, participants went through the following steps:
1) A landing page first explained the payment scheme; the classification task was presented (here, predicting the sentiment of reviews); and they were shown dataset-specific examples. 
2) To familiarize them with the task and verify their understanding, a screening phase required the participant to correctly label four of six reviews~\cite{liu-naacl16}.
Only participants who passed the gating test were allowed to proceed to the main task.
3) The main task randomly assigned participants to one of our study conditions (Section~\ref{sec:conditions_sentiment}) and presented condition-specific instructions, including the meaning and positioning of AI's prediction, confidence, and explanations.
Participants then labeled all 50 study samples one-by-one. 
For a given dataset, all conditions used the same ordering of examples.
\chifinal{The participants received immediate feedback on their correctness after each round of the task.}
4) A post-task survey was administered, asking whether they found the model assistance to be helpful, their rating of the usefulness of explanations in particular (if they were present), and their strategy for using model assistance.

We recruited participants from Amazon's Mechanical Turk, limiting the pool to subjects from within the United States with a prior task approval rating of at least 97\% and a minimum of 1,000 approved tasks.
To ensure data quality, we removed data from participants whose median labeling time was less than 2 seconds or those who assigned the same label to all examples. 
In total, we recruited 566 (\dbeer) and 552 (\damz) crowd workers, and in both datasets, 84\% of participants passed the screening and post-filtering. 
Eventually, we collected data from around 100 participants (ranging from 93 to 101 due to filtering) per condition.

Study participants received a base pay of \$0.50 for participating,
a performance-based bonus for the main task, and a fixed bonus of \$0.25 for completing the survey. 
Our performance-based bonus was a combination of linear and step functions on accuracy: we gave \$0.05 for every correct decision in addition to an extra \$0.50 if the total accuracy exceeded 90\% or \$1.00 if it exceeded 95\%. The assigned additional bonuses were intended to motivate workers to strive for performance in the complementary zone and improve over the AI-only performance~\cite{Ho2015IncentivizingHQ}.
Since we fixed the AI performance at 84\%, humans could not obtain the bonus by \checkchi{blindly} following the AI's recommendations.
Participants spent 13 minutes on average on the experiment and received an average payment of \$3.35 (equivalent to an hourly wage of \$15.77).

\paragraph{\noindent {\bf Modifications for LSAT}}
For the \dlsat dataset, we used a very similar procedure but used two screening questions and required workers to answer both correctly. We used a stricter passing requirement to avoid low-quality workers who might cheat, which we observed more for this task in our pilots. 
We again used MTurk with the same filters as sentiment classification, and we post hoc removed data from participants whose median response time was less than three seconds.
508 crowd workers participated in our study, 35\% of whom passed the screening and completed the main task, resulting in a total of 100 participants per condition.

Participants received a base pay of \$0.50 for participating, a performance-based bonus of \$0.30 for each correct answer in the main task, and a fixed bonus of \$0.25 for completing an exit survey. 
They received an additional bonus of \$1.00, \$2.00, and \$3.00 for reaching an overall accuracy of 30\%, 50\%, and 85\% to motivate workers to answer more questions correctly and perform their best.
The average completion time for the LSAT task was 16 minutes, with an average payment of \$6.30 (equals an hourly wage of \$23.34).

\begin{figure*}[ht]
 \centering
 \includegraphics[width=0.9\textwidth]{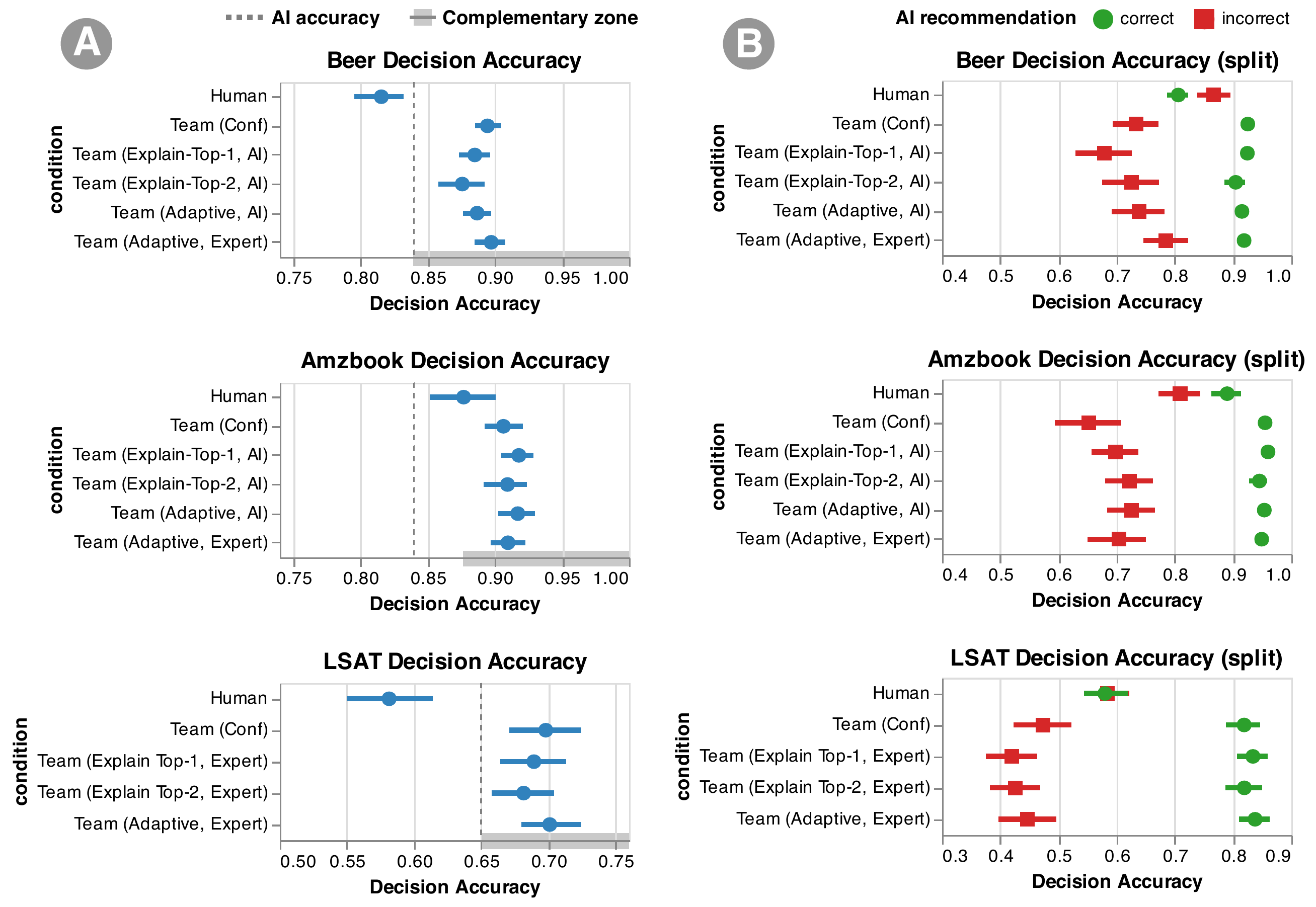}
 \vspace{-10pt}
 \caption{
 Team performance (with average accuracy and 95\% confidence interval) achieved by different explanation conditions and baselines for three datasets, with around 100 participants per condition.
 (A) Across every dataset, all team conditions achieved complementary performance.
 However, we did not observe significant improvements from using explanations over simply showing confidence scores. 
 (B) Splitting the analysis based on the correctness of AI accuracy, we saw that for \dbeer and \dlsat, \esingle explanations worsened performance when the AI was incorrect, the impact of \esingle and \edouble explanations were correlated with the correctness of the AI's recommendation, and \eadapt explanations seemed to have the potential to improve \esingle when the AI was incorrect, and to retain the higher performance of \esingle when the AI was correct.
}
\Description{
This figure summarizes the team performance achieved by different explanation conditions and baselines for three datasets, with around 100 participants per condition.
Across every dataset, all team conditions achieved complementary performance.
However, we did not observe significant improvements from using explanations over simply showing confidence scores. 
}
\vspace{-5pt}
 \label{fig:result}
\end{figure*}

\section{Results}
\label{sec:result}

\subsection{Effect of Explanation on Team performance}
\label{subsec:result_quan}
Figure~\ref{fig:result}A shows the team performance (\ie, accuracy of final decision) for each domain and condition. We tested the significance of our results using Student's T-tests with Bonferroni correction.

{\bf The baseline team condition, \cConf, achieved complementary performance across tasks.}
For \dbeer, providing AI recommendations and confidence to users increased their performance to ($\mu=0.89 \pm \sigma=0.05$), surpassing both AI ($0.84$) and unassisted human accuracy ($0.82 \pm 0.09$). Similarly, \cConf achieved complementary performance for \damz and \dlsat, with relative gains of 2.2\% and 20.1\% over unassisted workers (Figure~\ref{fig:result}A). 

{\bf We did not observe a significant difference between \esingle and \edouble}, or that \ref{h:single_double} was not supported. 
For example, in Figure~\ref{fig:result}A of \dbeer, explaining the top prediction performed marginally better than explaining the top-two predictions, but the difference was not significant ($z$=0.85, $p$=.40). 
The same was true for \damz ($z$=0.81, $p$=.42) and \dlsat ($z$=0.42, $p$=.68).

{\bf We did not observe significant improvements over the confidence baseline by displaying explanations.}
For example, for \dbeer, \cConf and \cAISingle achieved similar performance, with the accuracy being $0.89 \pm 0.05$ vs. $0.88 \pm 0.06$ respectively; the difference was insignificant ($z$=-1.18, $p$=.24). We observed the same pattern for \damz ($z$=1.23, $p$=.22) and \dlsat ($z$=0.427, $p$=.64).
As a result, we could not reject our hypothesis \ref{h:conf} that \esingle performs similar to simply showing confidence . 
This result motivates the need to develop new AI systems and explanation methods that provide true value to team performance by supplementing the model's confidence, perhaps working in tandem with confidence scores.

Though designed to alleviate the limitations of \esingle and \edouble in our experiments, {\bf we did not observe improvements from using \eadapt explanations.} 
For example, we did not observe any significant differences between \cAIAdapt and \cConf for \dbeer ($z$=-1.02, $p$=.31) or \damz ($z$=1.08, $p$=.28). We did not observe significant differences between \cExpertAdapt and \cConf for \dlsat ($z$=0.16, $p$=.87). 
More surprisingly, switching the source of \eadapt explanation to expert-generated did not significantly improve sentiment analysis results. For example, in Figure~\ref{fig:result}A, the differences in performance between \cExpertAdapt and \cAIAdapt were insignificant: \dbeer ($z$=1.31, $p$=.19) and \damz ($z$=-0.78, $p$=.43). As such, we could not reject the null hypotheses for either \ref{h:present} or \ref{h:quality}.

\begin{figure*}
 \centering
 \includegraphics[width=0.78\textwidth]{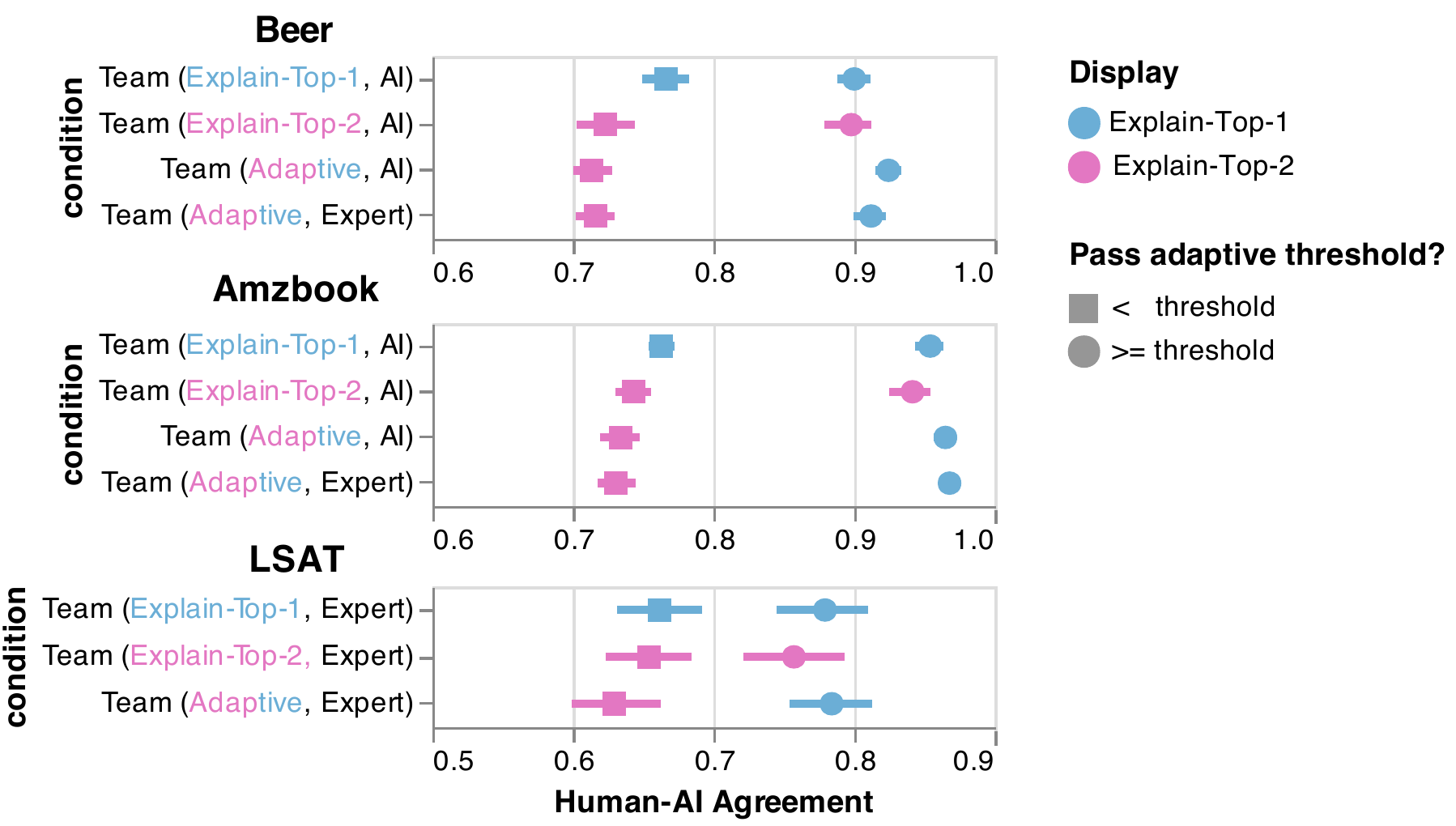}
 \vspace{-5pt}
 \caption{
 Relative agreement rates between humans and AI (\ie, does the final human decision match the AI's suggestion?) for various conditions, with examples split by whether AI's confidence exceeded the threshold used for \eadapt explanations.
 Across the three datasets, \eadapt explanations successfully reduced the human's tendency to blindly trust the AI \checkchi{(\ie, decreased agreement)} when it was uncertain and more likely to be incorrect. 
 For example, comparing \cAISingle and \cAIAdapt on low confidence examples that did not pass the threshold (rectangles), participants in \edouble (pink rectangles) were less likely to agree with the AI compared to those who saw \esingle (blue rectangles). 
}
\Description{
This figure shows the relative agreement rates between humans and AI (or, does the final human decision match the AI's suggestion?) for various conditions, with examples split by whether AI's confidence exceeded the threshold used for Adaptive explanations.
 Across the three datasets, adaptive explanations successfully reduced the human's tendency to blindly trust the AI when it was uncertain and more likely to be incorrect. 
}
 \label{fig:result_split_adapt}
\end{figure*}

{\bf While \eadapt explanation did not significantly improve team performance across domains, further analysis may point a way forward by combining the strengths of \esingle and \edouble.} Split the team performance by whether the AI made a mistake (Figure~\ref{fig:result}B), we observe that explaining the top prediction lead to better accuracy when the AI recommendation was correct but worse when the AI was incorrect, as in our pilot study. 
This is consistent with Psychology literature~\cite{koehler-pb91}, which has shown that human explanations cause listeners to agree even when the explanation is wrong\chifinal{, and recent studies that showed explanations can mislead data scientists into overtrusting ML models for deployment~\cite{kaur-chi2020}. While these results were obtained by measuring user's subjective ratings of trust, to the best of our knowledge, our studies are the first to show this phenomenon for explanation and end-to-end decision making with large-scale studies.}
As expected, in \dbeer, \eadapt explanations improved performance over \esingle when the AI was incorrect and improved performance over \edouble when the AI was correct, although the effect was smaller on other datasets.

\begin{figure*}[t]
	\centering
	\includegraphics[width=0.9\textwidth]{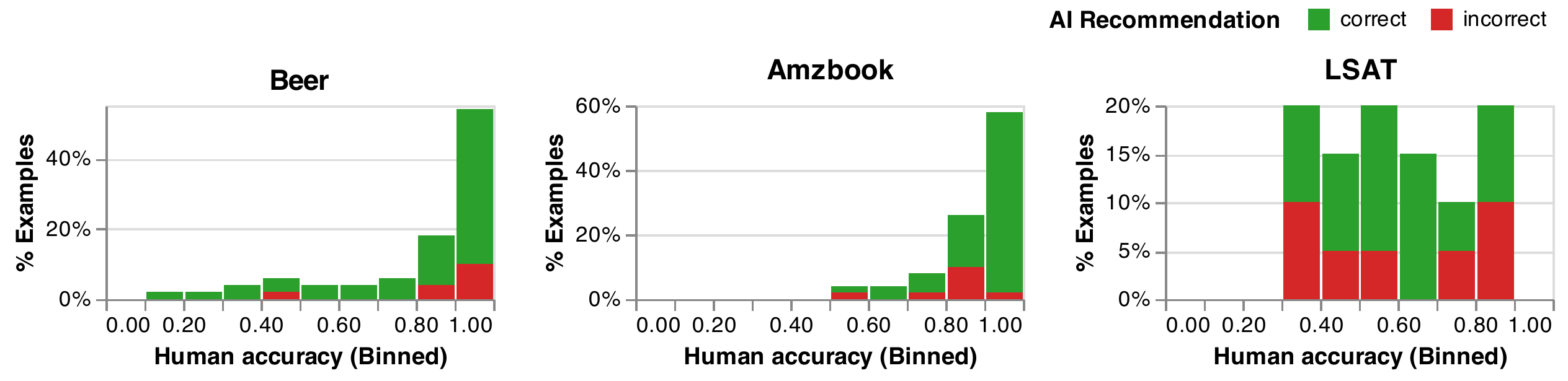}
	\vspace{-5pt}
	\caption{
 The distribution of study examples as a function of average human accuracy. For each domain, examples on the right were easy for most humans working alone. 
 Both \dbeer and \dlsat show a distribution that shows potential for complementary team performance: humans can correct easy questions mistaken by the AI (red bars towards the right), and, conversely, the AI may add value on examples where humans frequently err (green bars towards the left).
 In contrast, \damz showed less potential for this kind of human-AI synergy, with less ``easy for human'' questions (bars towards the left).
 }
 \Description{
  The distribution of study examples as a function of average human accuracy. For each domain, examples on the right were easy for most humans working alone. 
 Both Beer and LSAT show a distribution that shows potential for complementary team performance: humans can correct easy questions mistaken by the AI (red bars towards the right), and, conversely, the AI may add value on examples where humans frequently err (green bars towards the left).
 In contrast, AmzBook showed less potential for this kind of human-AI synergy, with less ``easy for human'' questions (bars towards the left).
 }
 \vspace{-5pt}
 \label{fig:complementary}
\end{figure*}

\begin{figure*}
\centering
\includegraphics[width=0.75\textwidth]{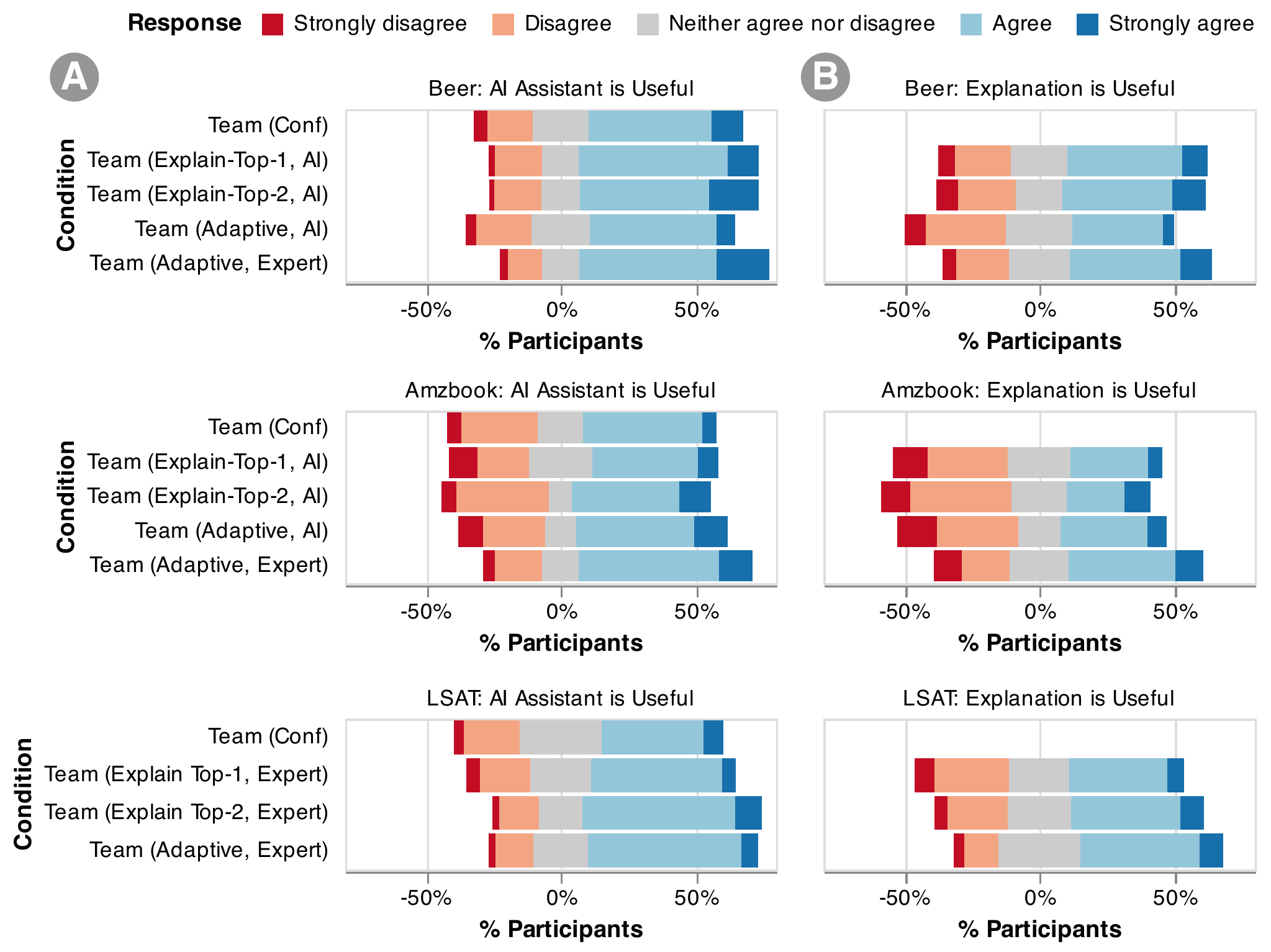}
\vspace{-5pt}
\caption{
Analysis of participant responses to two statements: 
(A) ``AI's assistance (\eg, the information it displayed) helped me solve the task'', and 
(B) ``AI's explanations in particular helped me solve the task.'' 
Across datasets, a majority of participants found AI assistant to be useful, and they rated all the conditions similarly, with a slight preference towards \cExpertAdapt.
In contrast to AI's overall usefulness, fewer participants rated explanations as useful, particularly \edouble explanations.
Participants also had a clearer preference for higher-quality (expert) \eadapt explanations.}
\Description{
Analysis of participant responses to two statements: 
(A) ``AI's assistance helped me solve the task'', and 
(B) ``AI's explanations in particular helped me solve the task.'' 
Across datasets, a majority of participants found AI assistant to be useful, and they rated all the conditions similarly, with a slight preference towards Expert explanations.
In contrast to AI's overall usefulness, fewer participants rated explanations as useful, particularly Double explanations across sentiment classification datasets.
Participants also had a clearer preference for higher-quality (expert) Adaptive explanations.}
\vspace{-5pt}
\label{fig:survey}
\end{figure*}

While Figure~\ref{fig:result}B shows team performance, the promising effects of \eadapt explanations are clearer if we study the agreement between AI predictions and human decisions (Figure~\ref{fig:result_split_adapt}). \eadapt explanations seem to encourage participants to consider the AI more when it is confident and solve the task themselves otherwise. Unfortunately, as our experiments show, the effect of using \eadapt did not seem sufficient to increase the final team accuracy, possibly for two reasons:
(1) in high confidence regions (circles in Figure~\ref{fig:result_split_adapt}), not only did workers have to agree more, but they also had to identify cases where the model failed with very high confidence (unknown unknowns~\cite{lakkaraju2017identifying}).
Identifying unknown unknowns could have been a difficult and time-consuming task for workers, and they may have needed other types of support that we did not provide.
(2) In low confidence regions (rectangles), not only did workers have to disagree more, but they also had to be able to solve the task correctly when they disagreed. 
\edouble explanations might have enabled them to suspect the model more, but it is unclear if they helped participants make the right decisions.
This indicates that more sophisticated strategies are needed to support humans in both situations.
We discuss some potential strategies in Section~\ref{subsec:when_discuss}.

{\bf \checkchi{Differences in expertise between human and AI affects whether (or how much) AI assistance will help achieve complementary performance.}}
To understand how differences in expertise between the human and AI impact team performance, we computed the average accuracy of unassisted users on study examples and overlaid the AI's expertise (whether the recommendation was correct) in Figure~\ref{fig:complementary}. 
The figure helps explain why users benefited more from AI recommendations for both \dbeer and \dlsat datasets.
There was a significant fraction of examples that the AI predicted correctly but humans struggled with (green bars to the left), while the same was not true for \damz (where AI recommendations did not help as much).
Further, when the AI was incorrect, explaining predictions on \damz via \esingle improved the performance by 5\% over showing confidence (Figure~\ref{fig:result}B), but it decreased the performance for \dbeer and \dlsat.
One possible explanation is that most AI mistakes were predicted correctly by most humans on \damz (red bars were mostly towards the right). 
After observing clear model mistakes, participants may have learned to rely on them less, despite the convincing-effect of explanation. Participants' self-reported collaboration approaches supported our guess -- \damz participants reportedly ignored the AI's assistance the most (Section~\ref{subsec:coding}). 
\chifinal{That said, other confounding effects such as the nature of the task (\eg, binary classification vs. choosing between multiple options) should also be studied.}

\renewcommand{\arraystretch}{1.1}
\newcommand{\surveycode}[1]{\emph{\texttt{#1}}\xspace}
\newcommand{\qFollowAI}{\surveycode{Mostly Follow AI}}
\newcommand{\qIgnoreAI}{\surveycode{Mostly Ignore AI}}
\newcommand{\qPreAI}{\surveycode{AI as Prior Guide}}
\newcommand{\qPostAI}{\surveycode{AI as Post Check}}

\newcommand{\qUseExpl}{\surveycode{Used Expl.}}
\newcommand{\qSpeedRead}{\surveycode{Speed Read}}
\newcommand{\qValidateAI}{\surveycode{Validate AI}}

\newcommand{\qUseConf}{\surveycode{Used Conf.}}
\newcommand{\qConfThreshold}{\surveycode{Conf. Threshold}}

\newcommand{\qBackAI}{\surveycode{Fall Back to AI}}
\newcommand{\qUpdate}{\surveycode{Updated Strategy}}

\begin{table*}
\centering
\setlength{\tabcolsep}{5pt}
\begin{tabular}{p{0.2\linewidth} p{0.5\linewidth} p{0.1\linewidth}}
\hline
\textbf{Codes} & \textbf{Definitions and Examples} & \textbf{\#Participants} \\
\toprule 
	\multicolumn{2}{l}{\textbf{Overall Collaboration Approach} (codes are mutually exclusive)}
	& \\
	\qFollowAI & 
	The participant mostly followed the AI.\newline
	\tquote{I went with Marvin most times.}
	& 23 (6\%)\\
	\qPreAI & 
	Used AI as a starting reference point.\newline
	\tquote{I looked at his prediction and then I read the passage.}
	& 190 (47\%)\\
	\qPostAI & 
	Double-checked after they made their own decisions.\newline
	\tquote{I ignored it until I made my decision and then verified what it said.}
	& 102 (25\%)\\
	\qIgnoreAI & 
	Mostly made their own decisions without the AI.\newline
	\tquote{I didn't. I figured out the paragraph for myself.}
	& 90 (22\%)\\
	\midrule
	\multicolumn{2}{l}{\textbf{The Usage of Explanation} (codes can overlap)} \\
	\qUseExpl & 
	Explicitly acknowledged they used the explanation.\newline
	\tquote{I skimmed his highlighted words.}
	& 138 (42\%)\\
	\qSpeedRead & 
	Used explanations to quickly skim through the example.\newline
	\tquote{I looked at Marvin's review initially then speed read the review. }
	& 29 (9\%)\\
	\qValidateAI & 
	Used the explanation to validate AI's reasoning.\newline
	\tquote{Marvin focuses on the wrong points at times. This made me \newline cautious when taking Marvin's advice.}
	& 17 (5\%)\\
	
	\midrule
	\multicolumn{2}{l}{\textbf{The Usage of Confidence} (codes can overlap)} \\
	\qUseConf & 
	Explicitly acknowledged they used the confidence.\newline
	\tquote{I mostly relied on Marvin's confident levels to guide me.}
	& 90 (22\%)\\
	\qConfThreshold & 
	Was more likely to accept AI above the threshold.\newline
	\tquote{If Marvin was above 85\% confidence, I took his word for it.}
	& 24 (6\%)\\
	
	\midrule
	\multicolumn{2}{l}{\textbf{Others} (codes can overlap)} \\
	\qBackAI & 
	Followed the AI's label if they failed to decide.\newline
	\tquote{I used it if I was unsure of my own decision.}
	& 54 (13\%)\\
	\qUpdate & 
	Changed their strategy as they interacted more.\newline
	\tquote{I decided myself after seeing that sometimes Marvin failed me. }
	& 12 (2\%)\\
\bottomrule
\end{tabular}
\caption{The codebook for participants' descriptions of how they used the AI, with the number of self-reports.}
\Description{The codebook for participants' descriptions of how they used the AI, with the number of self-reports.}
\vspace{-15pt}
\label{table:coding}
\end{table*}

\subsection{Survey Responses on Likert Scale Questions}
Two of the questions in our post-task survey requested categorical ratings of AI and explanation usefulness.\footnote{\chifinal{Since we did not pre-register hypotheses for the subjective ratings and only analyzed them post-hoc, we do not perform/claim statistical significant analysis on these metrics.}}

\textbf{AI usefulness}: While participants generally rated AI assistance useful (Figure~\ref{fig:survey}A), the improvements in ratings between most explanations and simply showing confidence were marginal. 
The difference was more clear for high-quality adaptive explanations; for \dbeer, 70\% of the participants rated AI assistance useful with \cExpertAdapt in contrast to 57\% with \cConf. 
We observed a similar pattern on \damz (65\% vs. 49\%) and \dlsat (63\% vs. 45\%), though on \dlsat, \cExpertDouble received slightly higher ratings than \cExpertAdapt (66\% vs. 63\%).

\textbf{Explanation usefulness}: Figure~\ref{fig:survey}B shows that participants' ratings for the usefulness of explanations were lower than the overall usefulness of AI's assistance (in A). 
Again, expert-generated \eadapt explanations received higher ratings than AI-generated ones for \dbeer(53\% vs. 38\%) vs. \damz (50\% vs. 40\%).
This could indicate that showing higher quality explanations improves users' perceived helpfulness of the system. 
However, it is worth noting that this increased preference did not translate to an improvement in team performance, which is consistent with observations made by Bu\c{c}inca \etal~\cite{buccinca-iui20} that show that people may prefer one explanation but make better decisions with another.


\label{sec:post_survey}

\subsection{Qualitative Analysis on Collaboration}
\label{subsec:coding}

\begin{figure*}[t]
 \centering
 \includegraphics[width=0.8\textwidth]{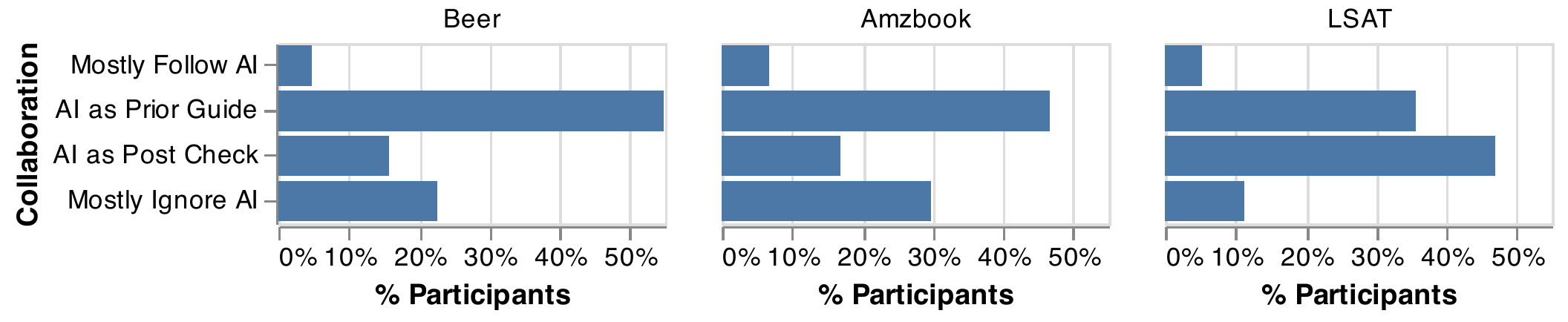}
 \vspace{-5pt}
 \caption{
Instead of ignoring or strictly following the AI, participants reported taking the AI information into consideration most of the time.
 They most frequently used AI as a prior guide in sentiment analysis, but used it as post-check in \dlsat.
 They were also more likely to ignore the AI in sentiment analysis than in \dlsat.
 }
 \Description{Instead of ignoring or strictly following the AI, participants reported taking the AI information into consideration most of the time.
 They most frequently used AI as a prior guide in sentiment analysis, but used it as post-check in LSAT.
 They were also more likely to ignore the AI in sentiment analysis than in LSAT.
}
 \label{fig:qual_strategy}
 \bigskip
 \includegraphics[width=0.85\textwidth]{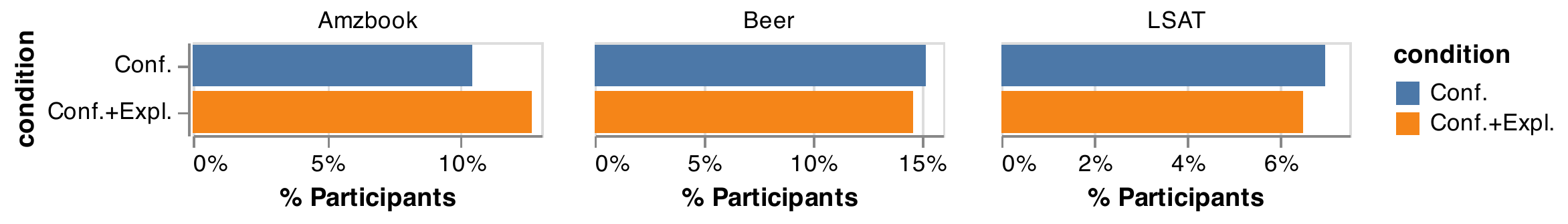}
 \vspace{-10pt}
 \caption{Comparing the occurrence of \emph{\qUseConf} in just the confidence condition and in those with explanations, we saw a similar proportion of users that explicitly acknowledged using confidence, regardless of whether they saw an explanation.}
  \Description{Comparing the occurrence of the code ``used conf'' in just the confidence condition and in those with explanations, we saw a similar proportion of users that explicitly acknowledged using confidence, regardless of whether they saw an explanation.
}
 \label{fig:conf_collected_from}
\end{figure*}

To better understand how users collaborated with the AI in different tasks, we coded their response to the prompt: ``Describe how you used the information Marvin (the AI) provided.'' 
Two annotators independently read a subset of the responses to identify emergent codes and, using a discussion period, created a codebook (Table~\ref{table:coding}).
Using this codebook, for each team condition and dataset, they coded a sample of 30 random worker responses: 28 were unique and 2 overlapped between annotators, allowing us to compute inter-annotator agreement.
Our final analysis used 409 unique responses after removing 11 responses deemed to be of poor quality (Table~\ref{table:coding}).
We scored the inter-annotator agreement with both the Cohen's $\kappa$ and the raw overlap between the coding.
We achieved reasonably high agreements, with an average $\mu(\kappa)=0.71, \sigma(\kappa)=0.18$ (the average agreement was $93\% \pm 6.5\%$).
We noticed the following, which echo the performance differences observed across datasets:

\textbf{Most participants used the AI's recommendation as a prior or to double-check their answers.}
For all datasets, more than 70\% of the participants mentioned they would partially take AI's recommendation into consideration rather than blindly following AI or fully ignoring it (Figure~\ref{fig:qual_strategy}).
Participants used the AI as a prior guide more than as a post-check for sentiment analysis, but not for \dlsat, which aligns with our interface design: for \dlsat, AI recommendations were on a separate pane, encouraging users to solve the task on their own before consulting the AI.

\textbf{Participants ignored the AI more on domains where AI expertise did not supplement their expertise.}
Figure~\ref{fig:qual_strategy} shows that while only 11\% of \dlsat participants claimed that they mostly ignored the AI, the ratio doubled (\dbeer, 23\%) or even tripled (\damz, 30\%) for sentiment analysis.
As discussed in Figure~\ref{fig:complementary}, this may be due to correlation differences between human and AI errors for different datasets: \damz participants were less likely to see cases where AI was more correct than they were, and therefore they may have learned to rely less on it. 
For example, one participant in \damz mentioned, 
\squote{I had initially tried to take Marvin's advice into account for a few rounds, and stopped after I got 2 incorrect answers. After that I read all of the reviews carefully and followed my own discretion.}

In contrast, a \dbeer participant relied more on the AI once realizing it could be correct:
\squote{At first I tried reading the passages and making my own judgments, but then I got several items wrong. After that, I just switched to going with Marvin's recommendation every time.}

In addition to the user's collaboration behavior, these differences between domains may have affected our quantitative observations of team performance. 
For example, a small difference between human and AI expertise (distribution of errors) means that the improvement in performance when the AI is correct would be less substantial. 
In fact, in Figure~\ref{fig:result}B, if we compare the team performance when the AI is correct, the difference between team conditions and the human baseline is least substantial for \damz.

\textbf{Some participants developed mental models of the AI's confidence score to determine {\em when} to trust the AI.}
Among participants who mentioned they used confidence scores (90 in total), 27\% reported using an explicit confidence threshold, below which they were likely to distrust the AI. 
The threshold mostly varied between 80 to 100 ($83\pm8$ for \dbeer, $89\pm 7$ for \damz, and $90 \pm 0$ for \dlsat) but could go as low as 65, indicating that users built different mental models about when they considered AI to be ``trustworthy.''
\chifinal{While this observation empirically shows that end-users develop mental model of trust in AI-assisted decision making~\cite{bansal-hcomp19}, it more importantly shows how the AI's confidence is a simple, yet salient feature via which users create a mental model of the AI's global behavior~\cite{gero-chi2020}.}
Note that across all three domains, the same proportion of participants self-reported using AI's confidence scores regardless of whether they saw explanations (Figure~\ref{fig:conf_collected_from}).
\\
Furthermore, \textbf{some participants consigned the task to AI when they were themselves uncertain.} For example, 13\% participants mentioned that they would go with the AI's decision if they were on the fence by themselves: \squote{There were some that I could go either way on, and I went with what Marvin suggested.}
\checknew{These user behaviors are similar to observations in psychology literature on {\em Truth-Default Theory}~\cite{levine-jlsp2014}, which shows that people exhibit {\em truth-default} behavior: by default, people are biased to assume that the speaker is being truthful, especially when {\em triggers} that raise suspicion are absent. Furthermore, our participants' distrust in low-confidence recommendations is also consistent with examples of triggers that cause people to abandon the truth-default behavior.}


\textbf{Explanations can help participants validate the AI's decisions, and the inline-highlight format helped participants speed up their decision making.}
Among the participants who explicitly mentioned using explanations, 27\% in \dbeer and 32\% in \damz reported that they used them to read the review text faster.
Since \dlsat explanations required reading additional text, we did not expect \dlsat users to find this benefit.
Interestingly, for \dbeer and \damz, while a small percentage of users (17\%) reported using the explanations to validate the AI's decisions (see Figure~\ref{table:coding}), only 2\% did so in \dlsat. This could be because \dlsat is a harder task than sentiment analysis, and verifying AI's explanations is costlier.
Other participants mostly mentioned that they would supplement their own reasoning with the AI's:
\squote{I read the Marvin rationale and weighed it against my intuition and understanding.}

\section{Discussion \&\ Future Directions}
\label{sec:discuss}
Though conducted in a limited scope, our findings should help guide future work on explanations and other mechanisms for improving decision making with human-AI teams. 

\subsection{Limitations}
\label{subsec:limitations}

As mentioned in Section~\ref{sec:background}, AI explanations have other motivations not addressed by this paper.
Our work, as well as the papers listed in Table~\ref{table:prior_work}, evaluated team performance along one dimension: accuracy of decisions. 
We did not explore the benefits on other metrics (\eg increasing speed as reported by some users in Section \ref{sec:post_survey}), but in general, one may wish to achieve complementary performance on a multi-dimensional metric.
\checknew{In fact, research shows that large collaborative communities like Wikipedia require AI systems that balance multiple aspects, \eg, reducing human effort, improving trust and positive engagement~\cite{smith-cscw2020}.} 
We encourage future research to extend the definition of complementarity, and to evaluate the impact of explanations on those dimensions accordingly.

\chifinal{
Further, we restricted ourselves to tasks amenable to crowdsourcing (text classification and question answering), so our results may not generalize to high-stakes domains with expert users such as medical diagnosis. 
We also note that the effectiveness of explanations may depend on user expertise, a factor that we did not explore.
Investigating this in our framework would either require recruiting lay and expert users for the same task~\cite{feng-iui19} or utilizing a within-subject experimental design to measure user expertise.
}

Finally, we only explored two possible ways to present explanations (highlighting keywords and natural language arguments). While these methods are widely adopted~\cite{wang-etal-2016-attention,Lin2017ASS,Ghaeini-abs-1808-03894,lai-fat19}, alternative approaches may provide more benefit to team performance.

\subsection{Explaining AI for Appropriate Reliance} 
One concerning observation was that explanations increased blind trust rather than appropriate reliance on AI.
This is problematic especially in domains where humans are required in the loop for moral or legal reasons (\eg, medical diagnosis) and suppose the presence of explanations simply soothes the experts (\eg, doctors), making them more compliant so they blindly (or become more likely to) agree with the computer. 
Encouraging human-AI interactions like these seems deeply unsatisfactory and ethically fraught.
\chifinal{
Importantly, while prior works also observed instances of inappropriate reliance on AI~\cite{wang2019designing, croskerry2009clinical, kaur-chi2020, mitchell2019model}, our studies quantified its effect on team performance.
Since the nature of the proxy tasks can significantly change the human behavior, they can lead to potential misleading conclusions~\cite{buccinca-iui20}. 
The emphasis of the complementary team performance in end-to-end tasks can \emph{objectively} evaluate the extent of such issues or about the effectiveness of a solution.}

\Comment{
\checkchi{While other metrics can be used for evaluation (more discussed in Section~\ref{subsec:limitations}), we directly evaluate end-to-end team accuracy for three reasons.}
First, deploying such a human-AI team is ideal if it achieves {\em complementary performance}, \ie, if it outperforms both the AI and the human acting alone.
\checkchi{Second, evaluating explanations using proxy tasks (such as whether humans can use it to guess the model's prediction) can lead to different, misleading conclusions for achieving best team performance than an end-to-end evaluation~\cite{buccinca-iui20}.} 
\checkchi{Third, } AI-assisted decision making is often listed as a major motivation for AI explanations. In recent years numerous papers have employed user studies to show that human accuracy increases if the AI system explains its reasoning for tasks as diverse as medical diagnosis, predicting loan defaults, and answering trivia questions.

Our result echoes prior work on inappropriate trust on systems~\cite{wang2019designing, croskerry2009clinical, kaur-chi2020, mitchell2019model}, \ie, explanations can lead humans to either follow incorrect AI suggestions or ignore the correct ones~\cite{bussone-ichi2015, stumpf2009interacting}.
However, using end-to-end studies, we go one step further to quantify the impact of such over-reliance on objective metrics of team performance.

}

Our \eadapt Explanation aims to encourage the human to think more carefully when the system had a low confidence.
\chifinal{While the relative agreement rates showed that the \edouble explanation might cue the humans to suspect the model's veracity (Figure~\ref{fig:result_split_adapt}), the method was not sufficient to significantly increase the final team accuracy (Figure~\ref{fig:result}).}
This is perhaps because users still have to identify high-confidence mistakes (unknown-unknowns) and solve the task when the AI is uncertain (Section~\ref{subsec:result_quan}).
A followup question is, then, what kind of interactions would help humans perform correctly when the AI is incorrect?

\paragraph{Explanations should be informative, instead of just convincing.}
Our current expert explanations did not help any more than the AI explanations, which may indicate that having the ML produce the {\em maximally convincing} explanation --- a common objective shared in the design of many AI explanation algorithms --- might be a poor choice for complementary performance~\cite{buccinca-iui20}.
\chifinaldel{Indeed, while it is known that human explainers and listeners often have different objectives~\cite{lombrozo-cogpsy07}, it seems quite problematic to develop AI systems that actively seek to minimize human oversight. }
A more ideal goal is explanations that accurately \emph{inform} the user -- such that the user can quickly gauge through the explanation when the AI's reasoning is correct and when it should raise suspicion.
\chifinaldel{We thought that this might happen with our RoBERTa-based explanations for sentiment analysis, but informal analysis showed that the AI usually found some plausible justification when it made mistakes.}
A successful example of this was seen with Generalized Additive Models (GAMs) for healthcare, where its global explanations helped medical experts suspect that the model had learned incorrect, spurious correlations (\eg a history of asthma reduces the risk of dying from pneumonia~\cite{caruana-kdd15}).
We hope future research can produce explanations that better enable the human to effectively catch AI's mistakes, rather than finding plausible justifications when it erred.

\paragraph{High complementary performance may require adapting beyond confidence.}
Since approaches based on confidence scores make it difficult to spot unknown-unknowns, instead it may be worthwhile to design explanation strategies that adapt based on the {\em frequency} of agreement between the human and AI. 
For example, instead of explaining why it believes an answer to be true, the AI might play a devil's advocate role, explaining its doubts --- even when it agrees with the human.
The doubts can even be expressed in an interactive fashion (as a back and forth conversation) than a set of static justifications, so to avoid cognitive overload. For example, even if the system agrees with the user, the system can present a high-level summary of evidence for top-K alternatives and let the user drill down, \ie, ask the system for more detailed evidence for the subset of alternatives that they now think are worth investigating.

\subsection{Rethinking AI's Role in Human-AI Teams}
\label{subsec:when_discuss}

\paragraph{\checknew{Comparable accuracy does not guarantee complementary partners.}}

Rather, in an ideal team, the human and AI would have minimally overlapping mistakes so that there is a greater opportunity to correct each other's mistakes.
In one of our experiment domains (\damz), AI errors correlated much more strongly with humans' than in others, and thus we saw relatively smaller gains in performance from AI assistance (Figure~\ref{fig:complementary}).
As recent work has suggested~\cite{Wilder2020LearningTC,madras-neurips18,mozannar-arxiv20,bansal2020OptimizingAF}, it may be useful to directly optimize for complementary behavior by accounting for the human behavior during training, who may have access to a different set of features~\cite{varshney-arxiv18}.

Furthermore, the human and AI could maximize their talents in different dimensions. For example, for grading exams, AI could use its computation power to quickly gather statistics and highlight commonly missed corner cases, whereas the human teacher could focus on ranking the intelligence of the student proposed algorithms~\cite{glassman2015overcode}.
Similarly, to maximize human performance at Quiz Bowl\chifinaldel{, which requires balancing both speed and accuracy}, Feng and Graber~\cite{feng-iui19} designed interaction so that the AI memorized and quickly retrieved documents relevant to a question, a talent which humans lacked because of cognitive limitations; however, they left the task of combining found evidence and logical reasoning to human partners. Future research should explore other ways to increase synergy.

\paragraph{The timing of AI recommendations is important.}
Besides the types of explanations, it is also important to carefully design {\em when} the AI provides its viewpoint.
All of our methods used a workflow that showed the AI's prediction (and its explanation) to the human, before they attempted to solve the problem on their own.
However, by presenting an answer and accompanying justification upfront, and perhaps overlaid right onto the instance, our design makes it almost impossible for the human to reason independently, ignoring the AI's opinion while considering the task. 
This approach risks invoking the anchor effect, studied in Psychology~\cite{englich2006playing} and \chifinal{introduced to the AI explanation field by Wang et al.~\cite{wang2019designing}} --- people rely heavily on the first information that is presented by others when making decisions.
This effect was reflected in an increase in the use of the ``\qPreAI'' collaboration approach in the sentiment analysis domain, compared to \dlsat (Figure~\ref{fig:qual_strategy}).

Alternate approaches that present AI recommendations in an asynchronous fashion might increase independence and improve accuracy.
For example, pairing humans with slower AIs (that \checkchi{wait or} take more time to make recommendation) may provide humans with a better chance to reflect on their own decisions~\cite{park2019slow}.
Methods that embody recommendations from management science for avoiding group-think~\cite{macleod2011avoiding} might also be effective, \eg, showing the AI's prediction after the human's initial answer or only having the AI present an explanation if it disagreed with the human's choice.
We note that these approaches correspond to the Update and Feedback methods of Green \& Chen~\cite{green-cscw2019}, which {\em were} effective, albeit not in the complementary zone. 
Another approach is to limit the AI's capabilities. For example, one might design the AI to summarize the best evidence for all possible options, without giving hard predictions, by training {\em evidence agents}~\cite{perez-emnlp20}.
However, by delaying display of the AI's recommendation until after the human has solved the task independently \checkchi{or restricting to only per class evidences}, one may preclude improvement to the {\em speed} of problem solving, which often correlates to the cost of performing the task.

As a result, there is a strong tension between the competing objectives of speed, accuracy, and independence; 
We encourage the field to design and conduct experiments and explore different architectures for balancing these factors.

\section{Conclusions}
Previous work has shown that the accuracy of a human-AI team can be improved when the AI explains its suggestions, but these results are only obtained in situations where the AI, operating independently, is better than either the human or the best human-AI team. We ask if AI explanations help achieve \emph{complementary} team performance, \ie whether the team is more accurate than either the AI or human acting independently. 
We conducted large-scale experiments with more than 1,500 participants. Importantly, we selected our study questions to ensure that our AI systems had accuracy comparable to humans and increased the opportunity for seeing complementary performance.
While all human-AI teams showed complementarity, none of the explanation conditions produced an accuracy significantly higher than the simple baseline of showing the AI's confidence --- in contrast to prior work. 
Explanations increased team performance when the system was correct, but they decreased the accuracy on examples when the system was wrong, making the net improvement minimal. 

By highlighting critical challenges, we hope this paper will serve as a ``Call to action'' for the HCI and AI communities: and AI communities.
In future work, characterize when human-AI collaboration can be beneficial (\ie, when both parties complement each other), developing explanation approaches and coordination strategies that result in a complementary team performance that exceeds what can be produced by simply showing AI's confidence, and communicate explanations to increase understanding rather than just to persuade.
At the highest level, we hope researchers can develop new interaction methods that increase complementary performance beyond having an AI telegraph its confidence. 
\begin{acks}
This material is based upon work supported by ONR grant N00014-18-1-2193, NSF RAPID grant 2040196, the University of Washington WRF/Cable Professorship, and the Allen Institute for Artificial Intelligence (AI2), and Microsoft Research.
The authors thank
Umang Bhatt,
Jim Chen,
Elena Glassman,
Walter Lasecki,
Qisheng Li,
Eunice Jun,
Sandy Kaplan,
Younghoon Kim,
Galen Weld,
 Amy Zhang,
 and anonymous reviewers for helpful discussions and comments.
\end{acks}

\balance
\bibliographystyle{ACM-Reference-Format}
\bibliography{main}

\typeout{get arXiv to do 4 passes: Label(s) may have changed. Rerun}
\end{document}